# Towards Safer Transportation: a self-supervised learning approach for traffic video deraining

**Shuya Zong**
Graduate Research Assistant, Center for Connected and Automated Transportation (CCAT), and Lyles School of Civil Engineering, Purdue University, West Lafayette, IN, 47907.
Email: szong@purdue.edu

**Sikai Chen\***
Visiting Assistant Professor, Center for Connected and Automated Transportation (CCAT), and Lyles School of Civil Engineering, Purdue University, West Lafayette, IN, 47907.
Email: chen1670@purdue.edu; and
Visiting Research Fellow, Robotics Institute, School of Computer Science, Carnegie Mellon University, Pittsburgh, PA, 15213.
Email: sikaichen@cmu.edu
ORCID #: 0000-0002-5931-5619
(Corresponding author)

**Samuel Labi**
Professor, Center for Connected and Automated Transportation (CCAT), and Lyles School of Civil Engineering, Purdue University, West Lafayette, IN, 47907.
Email: labi@purdue.edu
ORCID #: 0000-0001-9830-2071

## ABSTRACT

Video monitoring of traffic is useful for traffic management and control, traffic counting, and traffic law enforcement. However, traffic monitoring during inclement weather such as rain is a challenging task because video quality is corrupted by streaks of falling rain on the video image, and this hinders reliable characterization not only of the road environment but also of road-user behavior during such adverse weather events. This study proposes a two-stage self-supervised learning method to remove rain streaks in traffic videos. The first and second stages address intra- and inter-frame noise, respectively. The results indicated that the model exhibits satisfactory performance in terms of the image visual quality and the Peak Signal-Noise Ratio value.

## INTRODUCTION

Traffic video captured by surveillance cameras is an integral part of intelligent transportation monitoring systems. The reliability and timeliness of real-time traffic information provided by traffic surveillance video has major influence on the roadway efficiency and safety. In the current era, most urban street corners are equipped with mounted surveillance cameras that assist the road agency in monitoring the intersection and adjacent traffic lanes. These facilities are important because increasing urbanization and motorization continue to exacerbate the problem of urban congestion and traffic accidents. The World Health Organization (WHO) reports that annually, 1.25 million people are killed and more than 50 million people incur injuries on roadways (*1*). In the US alone, roadway traffic accidents impose a direct economic cost of $242 billion annually, representing 1.6% of the U.S. Gross Domestic Product (*2*).

The dire safety situation associated with the road traffic environment could be addressed if the road authorities are able to monitor the roadway traffic environment for purposes of road-use enforcement, pre-crash evaluation for safety studies, crash risk assessments, liability purposes (*3-6*) and roadway characterization in the prospective era of autonomous vehicle operations. Monitoring of the roadway environment for these purposes is typically done using video images. In ideal weather and traffic conditions devoid of phenomena that degrade the visual quality of the driving environment, high quality video images of the roadway environment are obtained. However, in reality, certain natural events and traffic conditions may impair the acquisition of good images and jeopardize the task of roadway environment monitoring. The most common of these conditions is inclement weather, where streaks of falling rain drops severely degrade the visual quality of images of the traffic environment. Roadway traffic environment monitoring during rainfall is a challenging task for the human eye, let alone a video camera. Further, in a video, the motion of fast-moving cars generates motion noise in images that degrade the visual quality of videos.

The degradation of images due to natural or anthropogenic conditions have been duly recognized in the research literature. It has been found that these conditions significantly degrade the performance of monitoring equipment that rely on image/video feature extraction techniques (*7-10*), including event detection (*11*), image registration (*12*), object detection (*13, 14*), tracking, and recognition, scene analysis (*15*). Therefore, developing a video deraining algorithm to generate clean traffic video offers significant potential for more efficient traffic environment characterization and therefore, increased roadway safety. In addition, in highly automated transportation where the safety of driving maneuvers hinge largely on the quality and reliability of information received from local sensors including videos, video deraining and denoising in general, offers great promise in terms of safety.

Video deraining can be treated as a task of recovering a clean video $V = \{I_1, I_2, \ldots, I_n\}$ from a noisy video $\hat{V} = V + eN$, where $eN = \{N_1, N_2, \ldots, N_n\}$ and $\{N_t\}_{t=1}^n$ is the rain noise associated with each frame. Unlike the deraining of single image features such as a photo, the deraining of multiple image features such as videos, addresses not only the spatial noise for each frame but also the inter-frame temporal noise caused by motion. Therefore, some image-based video denoising approaches, which rely solely on the denoising of each single frame, tend to yield unsatisfactory results. Recently, convolutional neural network (CNN)-based video deraining algorithms have attracted much attention due to the quality of their end products (*16-19*). In this past work, CNNs were trained to learn to utilize temporal information by addressing clues between adjacent frames. The end-to-end CNNs have demonstrated great feasibility and flexibility in the video denoising task because they can be trained to remove various types of noise from videos. However, most of the end-to-end models therefore require pairs of noisy-clean images of the same location which may be difficult to obtain. Recently, Lehtinen et al. proposed a self-supervised "Noise2Noise" model that yielded satisfactory image-denoising outcomes (*20*). Unlike conventional models that remove rain noisy by learning the noisy rainy layer, the Noise2Noise model uses noisy images as both input and targets during the training and encourages the model to learn the average result among various noisy pairs. Inspired by their approach, we seek, in this study, to extend the N2N model into video deraining using a two-stage model. In the first stage, spatial noise is reduced using a single-image based N2N method. In the second stage, a regular spatial-temporal denoising method is applied by adopting the images denoised in first stage, as targets. This approach is therefore described as a self-supervised video deraining method. The contribution of the present study is two-fold:

1. A deep-learning model that removes streaks of falling rain and motion noise from a video, to support roadway characterization by the vehicle operator or autonomous system, real-time traffic management by the road agency, investigation of during-rain crashes by law enforcement, and data acquisition for roadway safety studies by researchers.

2. A proposed two-stage self-supervised learning approach that eliminates the challenges of acquiring matching pairs of noisy-clean images for training.

## LITERATURE REVIEW

### Still-Image Deraining

The recent years have witnessed significant progress in image deraining algorithms. Current image-deraining approaches fall into two groups: prior based algorithms (*21-23*) and data-driven CNN models (*24-27*). In the prior based algorithms, priors are proposed to detect rain streaks in images. Brewer and Liu (*28*) proposed a strategy for detecting rain streaks by checking whether the region exhibits a short duration intensity spike. In Elad and Aharon's work (*29*), it was assumed that image signals hold a sparse decomposition over a redundant dictionary. Another researcher (*30*) formulated a correlation model to capture the dynamics of falling rain, where the average of non-rain temporal neighboring pixels could be utilized to estimate rain density once the rain region is detected.

CNN based deraining algorithms have received significant attention recently due to their success and performance in image deraining (*24-27*). In DnCNN proposed in (*31*), residual learning and batch normalization were implemented for image denoising. The DnCNN has demonstrated its flexibility for tasks including blind Gaussian denoising, JPEG deblocking and image inpainting. Extended from DnCNN, FFDNet was developed to handle spatially variant noise (*32*). Qian et al. (*33*) utilized attentive generative network by adversarial training to recover a clean image from a raindrop-degraded image. Their model first learns about raindrop regions and their surroundings, and then focus on such regions to assess the local consistency for rain removal tasks. Another researcher (*34*) introduced DerainNet that directly learns the mapping relationship between rain versus and clean image detail layers from data.

### Video Deraining

Compared to still images, videos feature a strong temporal redundancy along motion trajectories. This added information in the temporal dimension provides more information when recovering a pixel from noisy frames. On the other hand, it also creates an extra degree of complexity which could be difficult to address. The movement of objects introduces motion noise which could be in the form of ghost flickering. Therefore, videos suffer from both spatial noise that exist in each frame and temporal noise. In this context, motion estimation and compensation has been employed in a number of video denoising algorithms to help to improve their performance and temporal consistency (*35-37*). Niklaus et al. (*38*) incorporated a pre-computed optical flow as motion information with a frame interpolation CNN. Caballero et al. (*39*) developed a network which estimate the motion by itself for video super resolution. Removal of the multiscale rain streaks in video was accomplished using a multiscale convolutional sparse coding established in prior research (*40*). Furthermore, considering different types of rain, a hybrid rain model is offered to model both rain streaks and occlusions using a dynamic routing residue recurrent network (*41*). However, most of the learning-based models mentioned above learn a noise layer and then remove the noise layer away from the original frames. In our proposed approach, rather than learning an explicit noise layer, the clean frame itself is what is learned instead.

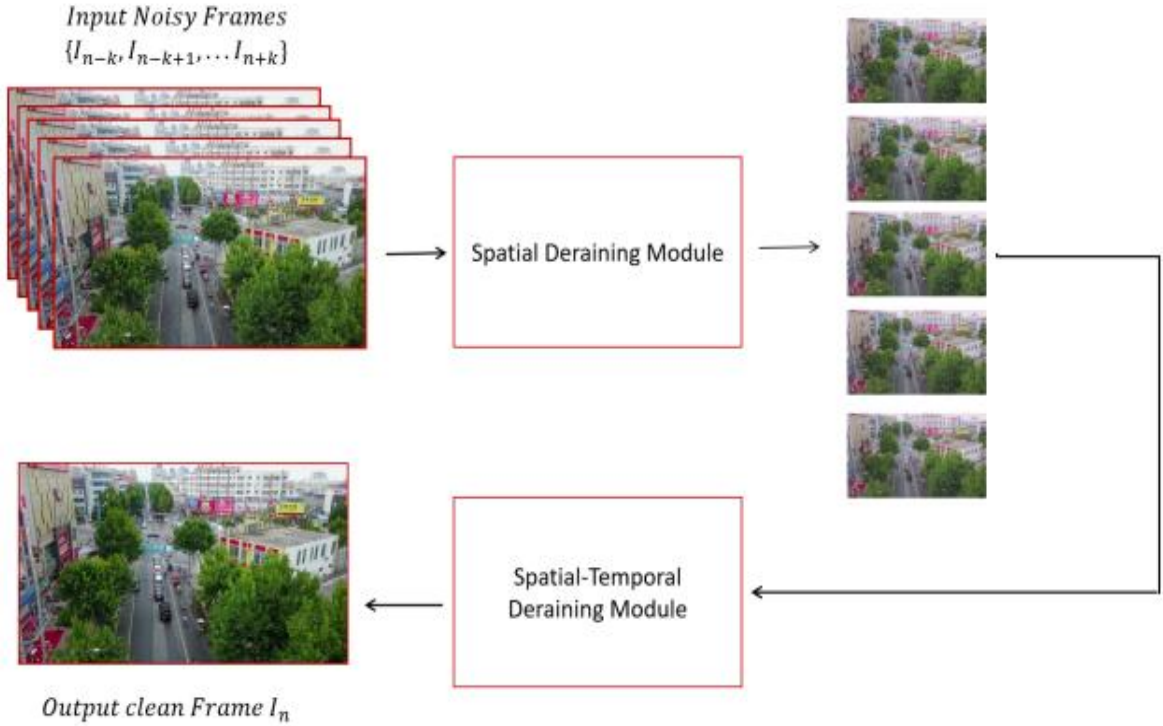

**Figure 1 The model structure**

## THE MODEL

For video denoising frameworks, flickering removal is crucial for improving the visual quality of the video. The flickering due to high-speed motion of objects or the motion of camera itself, could be removed using temporal clues that exist in neighborhood frames. Therefore, to recover a reference frame $I_n$, frames that neighbor $I_n$ are required. Consider a sequence of frames $\{I_{n-k}, I_{n-k+1}, \ldots, I_{n+k}\}$ in the neighborhood of the reference frame $I_n$ (*45*): the relationship between neighboring frames can be modeled as follows:

$$I_{n+k}(x) = I_n(x + \delta_k x) \; k \in \{-K, -K + 1, \ldots K\}$$

Where $\delta_k$ refers to the estimated optical flow between two frames. Several state-of-art video denoising models achieve a mapping function F that holds two main effects, namely motion estimation and prediction of the clean reference frame, which can be represented in Equation (1):

$$\hat{I}_n(x) = F(\{\hat{I}_{n+k}(x - \delta_k x)\}_{k=-K}^{K}) \qquad (1)$$

However, in all models mentioned above, clean-noisy pairs are required and the loss function is therefore expressed as:

$$L_{temp}(\theta) = \frac{1}{2m_t}\sum_{j=1}^{m_t} ||\hat{I}_j - I_j||_2 \qquad (2)$$

Where $\hat{I}_j$ is the estimated clean frame. Considering that acquiring such data is labor-demanding and challenging in real-word, we propose, in the present paper, a model that can be trained without noisy-clean pairs. Recently, one researcher (*41*) has verified, using FIVnet, that the "first image then video" approach adequately addresses the problem of blurred boundaries associated with moving objects and provides a state-of-art denoising result. In the FIVnet, intra-frame noise is firstly reduced seperately and then frames are wrapped together to remove inter-frame noises. Inspired by this concept in FIVnet, we herein propose a self-supervised video deraining model which consists of two stages: the first stage takes care of intra-frame noise while the second stage handles inter-frame noise. The deraining process can be expressed as follows:

$$\hat{I}_n(x) = \Phi(\{\varphi(I_{n+k}(x))\}_{k=-K}^{K})$$

Where $\varphi$ is a spatial deraining block and $\Phi$ is a spatial-temporal denoising block. Figure 1 presents the architecture of the model. In the spatial deraining module, we adopted Noise2Noise model which removes rain noise without requiring true labels. With regard to the spatial-temporal denoising module, we adopted a regular denoising model named FastDVDnet to take the derained images from $\varphi$ as input. This facilitated the task of removing motion noise, thus confirming the findings of past work (*42*). The details of each module are discussed in the sections that follow.

## The spatial deraining module

In the first module of our model, we implemented Noise2Noise (*20*) model to spatially derain 2K+1 frames. The Noise2Noise model provides an approach that can map observations corrupted by rain to clean images by learning mappings between noisy image pairs. In this model, neither an explicit statistical likelihood model of the noise nor a prior image is required. Therefore, unlike most existing noise removal models that remove noise by subtracting noise layers, Noise2Noise model learns the clean image itself. The theoretical background of this approach is that L2 loss learns to average observations. For example, it has been verified that in training the pairs of low-resolution and high-resolution images, L2 learns the average between all plausible explanations, which result in spatial blurriness to network output (*43,44*). Therefore, the authors of Noise2Noise model argued that one could add zero-mean noise to training and target sets without downgrading the network outputs. In the past research (*20*), different types of noise were tested for N2N but rain noise, specifically, was not considered. In this paper, we extend Noise2Noise to deraining tasks and use it as the spatial denoising module as the first part of our model. In the spatial denoising module, the loss function is:

$$L_N = \sum_{k=-K}^{K} ||\varphi(\hat{I}_{n+k}(x)) - \check{I}_{n+k}(x)||_2 \quad (3)$$

Where $\hat{I}_{n+k}(x)$ is noisy frames with rain noise i.i.d to the noise of $\check{I}_{n+k}(x)$.

## The spatial-temporal deraining module

In the spatial-temporal denoising stage, we used FastDVDnet (45) to further remove the noise due to object motion. As discussed in Section II, motion noise removal has been a challenging task in video processing. Conventional methods estimate the optical flow and fuse warped frames separately, while efforts have been made to promote the speed of video denosing by incorporating the two elements into one network by TOFlow (*46*). A more efficient network, namely FastDVDnet, is proposed as an alternative of motion estimation. As shown in figure 2, the FastDVDnest architecture contains two stages. The FastDVDnet processes five consecutive frames in the first denoising block and feeds the concatenated features into the second denoising block. Using this model, the explicit estimation of optical flow is skipped, which avoids the distortions and artifacts due to erroneous flow and speed up the training at the same time. Above all, the expensive computation of warping operations is also eliminated.

In our model, in order to avoid the FastDVDnet from learning the identity, we use different noise from spatial denoising module when training the spatial-temporal module. Instead of rain noise, Poisson noise is added when training FastDVDnet. Therefore, loss function of the second module is:

$$L_F = \sum_{n=1}^{N} ||\Phi(\hat{I}_n(x)) - \hat{I}_n(x)||_2 \quad (4)$$

Where $\hat{I}_n(x)$ is the estimated result from the first spatial denoising block and $\Phi$ represents the FastDVDnet model.

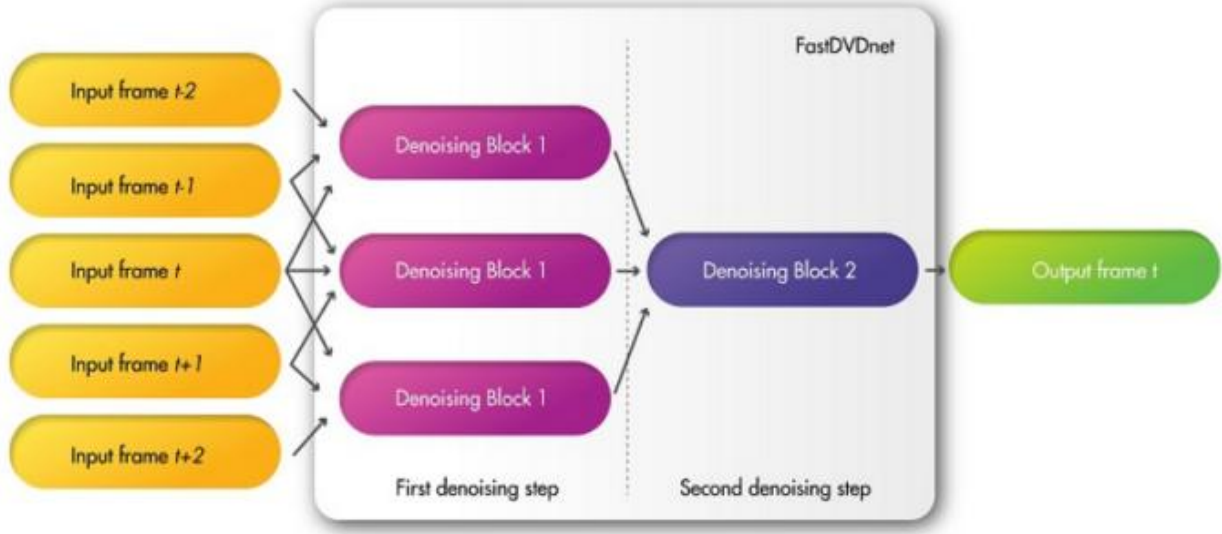

**Figure 2 Architecture of FastDVDnet (*45*)**

**Rain Noise Synthesis**

We synthesized rain noise for two reasons: First, the spatial draining module requires several rainy images of the same place. However, it is impossible to obtain such data in real-world since the real-world is quite dynamic. Even during consecutive rainy days, the configuration (e.g. location of vehicles, color of plants) of a scene is never the same. Secondly, in the absence of clean-rainy pairs of images, since we need to compare our deraining result against clean images for evaluation purposes, we synthesized rainy images from their clean versions, and used these for the analysis. Rain is a complicated atmospheric process that impairs the visual quality of a scene and also affects camera exposure time, depth of field and resolution. In most state-of-art deraining methods, the rain effect is assumed by specific models. To properly describe rain streaks, it is required to model rain drop size, rain density and rotation of rain due to wind. One way to carry out rain modelling is to carry out a linear superposition of the clean background and a layer of line shape rain streaks. A rainy scene $\tilde{R}$ can be modeled as:

$$\tilde{R}=R+B$$

where R is the clean background and B is the rain layer. In this study, we adopted transformed Gaussian noise, which is similar to the approach of PhotoShop software, to model the rain effect: the length and width of rain streak is added by noise stretching and the wind effect is modeled by rotating rain drops. Furthermore, rain also tends to introduce a mist effect to the video image. The accumulation and concentration of rain streaks forms an overall fog effect which renders the scene as a blurred image, to the human eye. Therefore, we also implemented a gaussian-blurring step to introduce blurring effect to the rainy scene. Figure 3 compares a clean image and its synthesized rainy version.

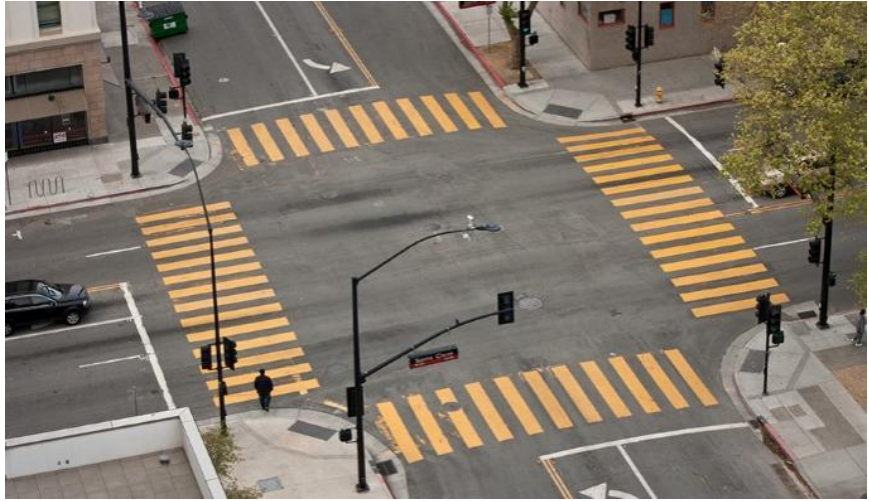

(a) Clean image (image during good weather)

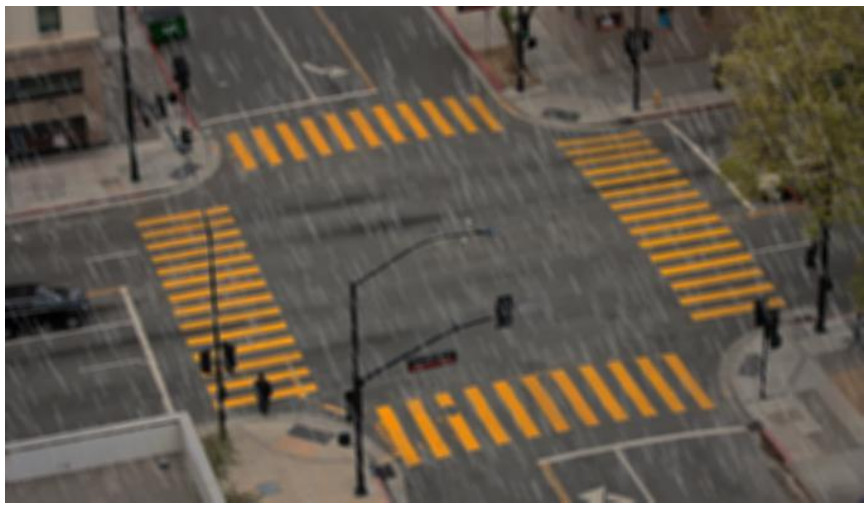

(b) Image during rain event

**Figure 3. Example of a clean image and a corrupted image**

## EXPERIMENT SETTING

We utilized the large-scale and high-quality Visdrone Dataset (*47*) in our experiments. This dataset provides 400 videos clips formed by 265,228 frames, captured by drone-mounted cameras, covering various real-world scenes. Similar to FastDVDnet, we realize the flow map estimation by the DeepFlow algorithm (*48*). During the training phase, we used 5 consecutive frames to recover 1 central frame and adopted 60 epochs for training. For the spatial deraining module, we used UNET (*49*) to extract features from video frames. Unet has been proved an effective network to obtain image features in many tasks including classification and detection. In the initial training of Noise2Noise model, rain noise with a mean density of 300 and standard deviation of 10 is added to the source images while the rain noise with a mean density of 500 and standard deviation of 20 is added to the target images.

To evaluate the deraining results, we calculated average Peak Signal-to-Noise Ratio (PSNR) values and Structural Similarity Index (SSIM) values between clean background and derained results for all sequences. For a video sequence, the PSNR and SSIM values are taken as the average score of each frame it holds. Given a reference image $f$ and a test image $g$ ($f$ and $g$ have the same size), the PSNR (dB) between $f$ and $g$ is defined by equations (5) and (6) below.

$$\text{PSNR}(f, g) = 10\log_{10}(255^2/\text{MSE}(f, g)) \qquad (5)$$

$$\text{where } \text{MSE}(f, g) = \frac{1}{MN}\sum_{i=1}^{M}\sum_{j=1}^{N}(f_{ij} - g_{ij})^2 \qquad (6)$$

Therefore, a higher value of PSNR indicates a superior deraining result. The Structural Similarity Index (SSIM) is a perceptual metric that quantifies the image quality degradation. It considers the structural information by the idea that pixels have strong inter-dependencies especially when they are spatially close. These dependencies carry important information about the structure of the objects in the visual scene. The SSIM of two images $x, y$ is calculated by:

$$SSIM = \frac{(2\mu_x\mu_y + c_1)(2\sigma_x\sigma_y + c_2)}{(\mu_x^2 + \mu_y^2 + c_1)(\sigma_x^2 + \sigma_y^2 + c_2)} \qquad (7)$$

Where

$\mu_x$ is the average of x;

$\mu_y$ is the average of y;

$\sigma_x$ is the variance of x;

$\sigma_y$ is the variance of y;

$c_1$ and $c_2$ are two variables to stablize the division;

L is the dynamic range of the pixel-values;

$k_1 = 0.01 \text{ and } k_2 = 0.03 \text{ by default}$

## RESULTS

### The deraining results

To demonstrate the effect of our model in traffic safety monitoring, both quantitative comparison and visual comparison are prepared. Table 1 quantitatively reports experimental results. Compared to rainy videos, our model achieves 16% and 10% improvement of PSNR and SSIM, respectively. We present a visualization of our model output in figure 4. The sub-figures in the top line are rainy scenes and the sub-figures at the bottom are derained results. The video sequences present a mix traffic flow formed by pedestrians and vehicles. It can be easily perceived that the result from our model have higher visual quality compared to the rainy scenes. It would be extremely difficult to count the number of people or observe the interactions between vehicles efficiently with rain-corrupted video frames. For instance, under the heavy rain, it is hard to tell how many pedestrians exist in the area we highlighted by a red circle. However, the existence of people and the behavior of both pedestrian and vehicles are captured clearly in the derained video frames. Therefore, by providing better visual qualities, our model can be utilized as a powerful tool for traffic monitoring and management.

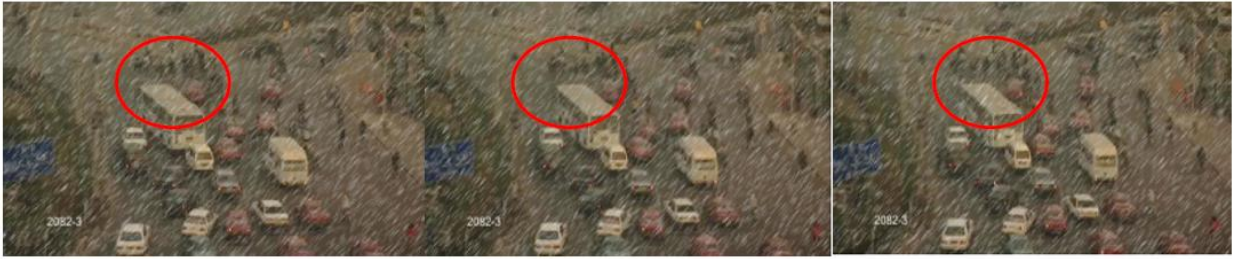

(a)  Video frames with rain noise

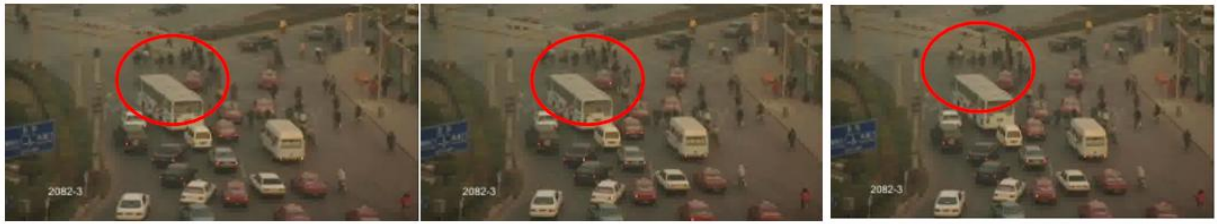

(b)  Video frames processed by our deraining model

**Figure 4. Deraining Results from Our Model**

**TABLE 1 Quantitative Results**

| Method | PSNR | SSIM |
|--------|------|------|
| Rainy Frames | 19 | 0.76 |
| Derained Results from our model | 22 | 0.89 |

## Comparisons and Discussions

To further explore the performance of the self-supervised video deraining approach, we conducted comparisons among our model and TOFlow (*46*) and GMM (*50*), which are existing flow-based denoising model and image-based denoising model, respectively. The image-based model is realized by processing each frame separately. Table 2 presents the difference of PSNR and SSIM values for derained results from different models. From the Table 2, it can be stated that even though no prior ground truth is required, we still reach comparable denoising performance as the supervised flow-based model TOFlow. Figure 5 presents a qualitative assessment of the visual outcomes from our model and the image-based model. It can be noticed that the problem of blur boundary is significant in image-based models in the area highlighted by red circles. This is because the temporal information is not taken into consideration when removing rain noise. Compared to image-based models, our model exhibits superior temporal coherence.

**TABLE 2 Quantitative Comparison (Rain density is the mean value for Gaussian rain noise)**

| Method | Rain Density of 500 | | Rain Density of 300 | |
|---|---|---|---|---|
| | PSNR | SSIM | PSNR | SSIM |
| Our Model | 18 | 0.87 | 22 | 0.89 |
| GMM (50) | 17 | 0.86 | 21 | 0.89 |
| TOFlow (46) | 19 | 0.90 | 22 | 0.92 |

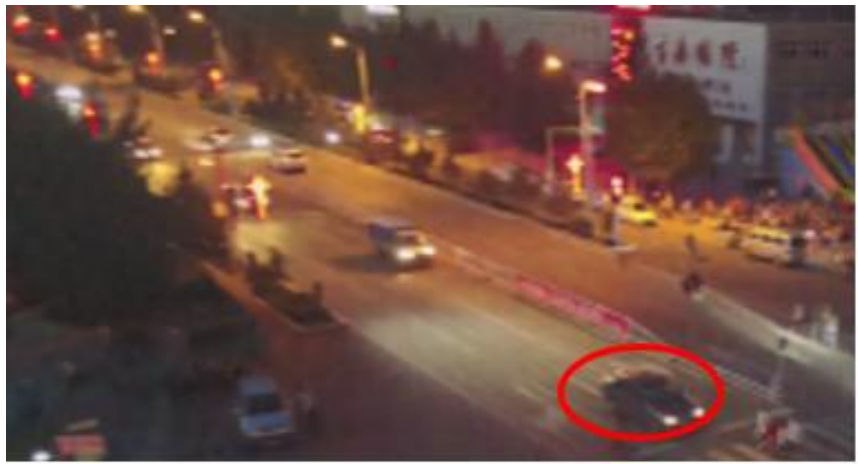

(a) Results from image-based model

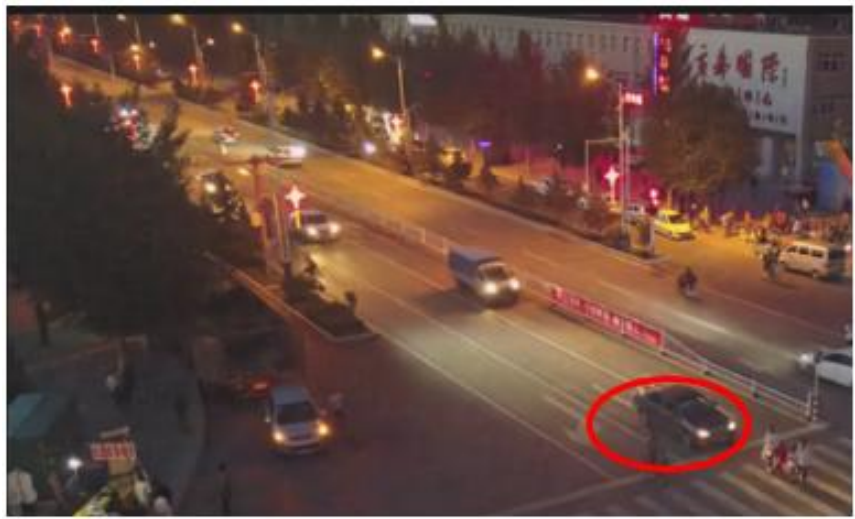

(b) Results from our model

**Figure 5. Comparison between results from the proposed model and the image-based deraining model**

In our proposed model, the two stages cooperate with (and benefit from) each other. Figure 6 presents an example of deraining result extracted from the first stage. It is observed that the derained image from the first stage is already clean, which significantly reduces the challenges for the second denoising phase. Consequently, the second denoising phase could focus on motion noise removal and give better outputs. However, when the original rain noise is heavy, the first deraining module may could not remove all noise.

Under this scenario, the following spatial-temporal denoising module could further remove the residual spatial noise. For instance, Figure 7 shows an example of the result from the first denoising module when the rain density is 800. In this figure, some rain streaks remain. However, as indicated by figure 8, after the spatial-temporal denoising phase, the rain streak noise became invisible. Table 3 gives the PSNR and SSIM values of images with and without the second stage and figure 9 presents the image quality difference in a more direct way. It can be observed that the second stage further improve video quality on the basis of the first stage, particularly when the rain noise is larger.

**TABLE 3 Results from two stages**

| | Results without spatial-temporal model (i.e. the second stage) | | Results with spatial-temporal model (i.e. the second stage) | |
|---|---|---|---|---|
| **Rain Density** | **PSNR** | **SSIM** | **PSNR** | **SSIM** |
| 300 | 21 | 0.89 | 22 | 0.89 |
| 500 | 16 | 0.85 | 18 | 0.87 |
| 800 | 15 | 0.83 | 17 | 0.86 |

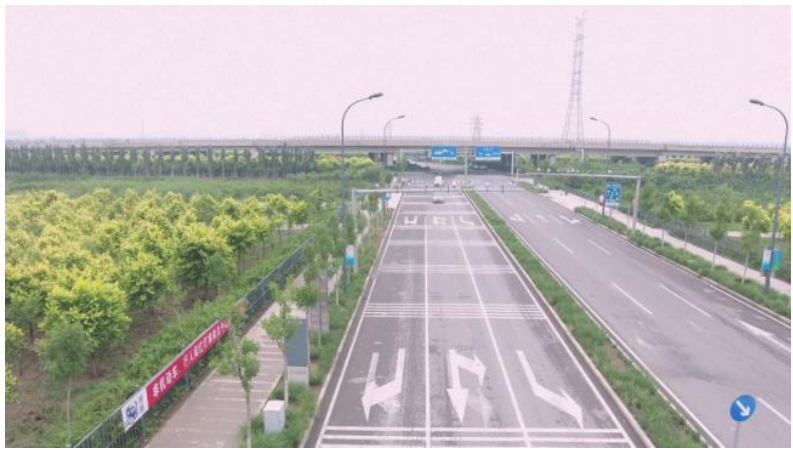

**Figure 6. Output from spatial deraining block 1**

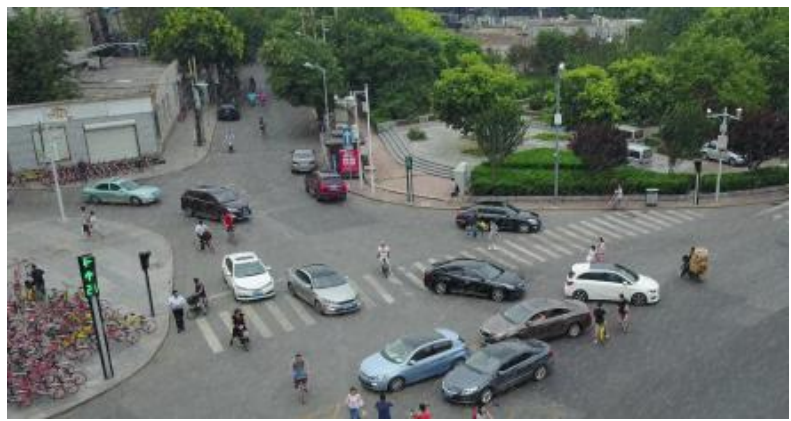

**Figure 7. Output from the spatial block may be not clean enough**

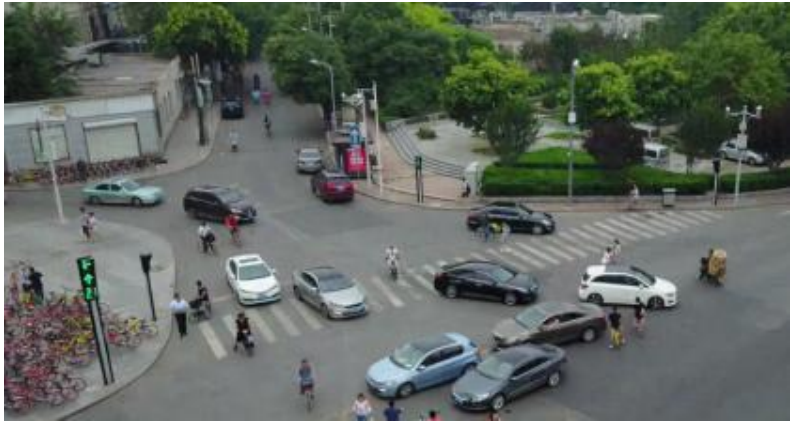

**Figure 8. Output from the spatial-temporal block**

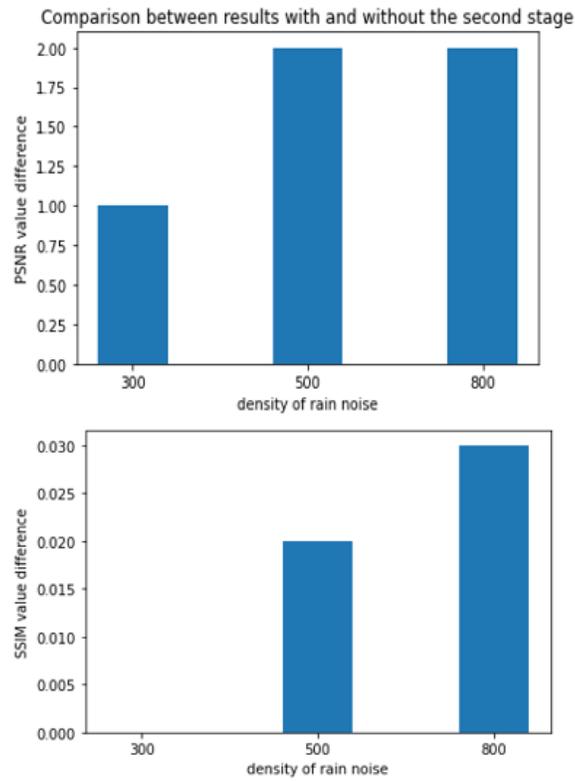

**Figure 9. Comparison between results with and without the second stage**

## CONCLUSION

In this work, we propose a two-stage model for video deraining to address the challenge of rain-induced blur in video images for purposes of enhanced traffic monitoring and management. In our two-stage model, both rain noise and motion noise are given full consideration. Rain noise is first reduced from each frame in the spatial denoising stage and then motion noise is removed by the following spatial-temporal denoising stage. A useful feature of our model is that it is self-supervised, requiring no ground-truth labels for video frames. Unlike previous work which were realized on the basis of noisy-clean training pairs, our model is built upon training through noisy-noisy pairs without much compromise in performance. Our experiments on Visdrone dataset have shown that although no ground truth is utilized in our model, it demonstrates performance comparable to that of supervised models. The self-supervised approach is achieved by executing the Noise2Noise model in the spatial denoising stage and then connecting it to FastDVDnet which serves as a spatial-temporal denoising module. The spatial-temporal denoising block strives to map corrupted video frames back to the results generated by spatial block instead of ground-truth clean frames. By removing rain noise from the scenes, the traffic surveillance video attains more pleasing visual quality, which could potentially improve the efficiency of video-supported roadway monitoring. The clean frames without noise not only demonstrate clearer, the behaviors of individual vehicles and pedestrians, but also provide visual friendly presentations of moving objects by removing the noised resulting from high-speed movement. However, there are several limitations of this work that lend opportunity for futher research in tis domain. First, we used synthesized rain rather than real rain. Although rain synthesis has been widely applied in deraining research, limited work has been done to verify whether the synthesized images capture the features of real-world rain. Secondly, we only considered rain in streak form. However, rain could also manifest in other shapes including circular rain drops which may cover a part of camera lens or rain mist that tend to span the entire environment space. It is expected that the requisite deraining methods will be different across these rain representations. In the future, our model could be tested on other types of rain representations and its performance assessed vis-à-vis other models. In addition, other self-supervised denoising models (e.g. self2void, self2self) could be investigated for their applicability to video deraining.

## ACKNOWLEDGMENTS
This work was supported by Purdue University's Center for Connected and Automated Transportation (CCAT), a part of the larger CCAT consortium, a USDOT Region 5 University Transportation Center funded by the U.S. Department of Transportation, Award #69A3551747105. The contents of this paper reflect the views of the authors, who are responsible for the facts and the accuracy of the data presented herein, and do not necessarily reflect the official views or policies of the sponsoring organization.
This manuscript is herein submitted for PRESENTATION ONLY at the 2022 Annual Meeting of the Transportation Research Board.

## AUTHOR CONTRIBUTIONS
The authors confirm contribution to the paper as follows: all authors contributed to all sections. All authors reviewed the results and approved the final version of the manuscript.